\documentclass[runningheads]{llncs}
\usepackage{tikz,pgfplots,graphicx,bm,floatrow,xcolor,todonotes}
\usepackage{tikz}
\usetikzlibrary{decorations.markings}
\tikzset{
    set arrow inside/.code={\pgfqkeys{/tikz/arrow inside}{#1}},
    set arrow inside={end/.initial=>, opt/.initial=},
    /pgf/decoration/Mark/.style={
        mark/.expanded=at position #1 with
        {
            \noexpand\arrow[\pgfkeysvalueof{/tikz/arrow inside/opt}]{\pgfkeysvalueof{/tikz/arrow inside/end}}
        }
    },
    arrow inside/.style 2 args={
        set arrow inside={#1},
        postaction={
            decorate,decoration={
                markings,Mark/.list={#2}
            }
        }
    },
}
\begin{document}
\title{Temporal Embeddings and Transformer Models for Narrative Text Understanding} 
\titlerunning{Temporal Embeddings and Transformers for Narrative Text Understanding}
\author{Vani K\ \and Simone Mellace \and Alessandro Antonucci}
\institute{Istituto Dalle Molle di Studi sull'Intelligenza Artificiale (IDSIA)\\Lugano (Switzerland)\\
\email{\{vanik,simone,alessandro\}@idsia.ch}}
\maketitle
\begin{abstract}
We present two deep learning approaches to narrative text understanding for character relationship modelling. The temporal evolution of these relations is described by \emph{dynamic word embeddings}, that are designed to learn semantic changes over time. An empirical analysis of the corresponding character trajectories shows that such approaches are effective in depicting dynamic evolution. A supervised learning approach based on the state-of-the-art transformer model BERT is used instead to detect static relations between characters. The empirical validation shows that such events (e.g., two characters belonging to the same family) might be spotted with good accuracy, even when using automatically annotated data. This provides a deeper understanding of narrative plots based on the identification of key facts. Standard clustering techniques are finally used for character \emph{de-aliasing}, a necessary pre-processing step for both approaches. Overall, deep learning models appear to be suitable for narrative text understanding, while also providing a challenging and unexploited benchmark for general natural language understanding.
\end{abstract}

\section{Introduction}\label{sec:intro}
Due to the inherent complexity involved in textual data, narrative text understanding remains a challenging and relatively unexplored research area for AI. Here we consider narrative text, such as novels and short stories (broadly termed here as \emph{literary text}) and try to address its lexical diversity and richness in terms of relations between entities \cite{piper2017studying}. In recent years, \emph{Deep Learning} (DL) approaches were found to positively impact \emph{Natural Language Processing} (NLP) with impressive boosts in text extraction and understanding capabilities. This marginally concerns the area of literary text \cite{labatut2019extraction}, where the application of DL models remains relatively unexplored. Some researchers modelled character networks using machine learning, mostly from a social network perspective based on generative models for conversational dialogues \cite{agarwal2012social,celikyilmaz2010actortopic} not involving DL state-of-the-art approaches. Just a few works have been reported for character evolution and relational analysis \cite{chaturvedi2016modeling,vani2019novel2graph,volpettitemporal}.

Here we evaluate the application of DL to literary text understanding. The goal is to describe character relationships within a novel and their evolution. Moreover, we also want to emphasize the potential of literary text as a challenging benchmark for state-of-the-art language models, whose major applications are typically in other domains such as biomedical literature 
\cite{li2017neural} or fake news detection \cite{ruchansky2017csi}, where both the lexical richness and the intricacy of the inter-entities relations might be less intricate compared to literary domain.

To analyse the character relationships, both supervised and unsupervised DL techniques are considered here. A classification model to identify the relations between characters using BERT (\emph{Bidirectional Encoder Representations from Transformers}, \cite{devlin2019bert}) is trained from supervised data. BERT is successfully used in various classification tasks, but, to the best of our knowledge, not yet in the literary text domain. Moreover, manual annotation of training data in this field can be very expensive, this representing a strong limitation for this direction. To partially bypass this issue, here we also present a simple approach to automatically generate training data for character relation classification (focusing on family relations, such as \emph{parent of}, \emph{sibling of}). 

At the unsupervised level, we consider the dynamic evolution of the characters over time (i.e., across the text). To do that, we learn vectors associated to different characters based on so-called \emph{dynamic or temporal embeddings} \cite{bamler2017dynamic}, allowing to learn vectors over different slices inside the text (e.g., chapters or fixed amounts of text), while maintaining the vectors comparable over time because of a common initialization. We analyse the relations between characters by visualizing the character trajectories over time by low-dimensionality projections or the relative distances in the original, high-dimensionality, spaces or by considering the relative distances between the vectors.

Both techniques require a pre-processing step consisting in character detection, based on standard entity recognition techniques, and character de-aliasing, for which density-based clustering methods are adopted.

The paper is organized as follows. A review of existing work is in Section \ref{sec:existing}. Sections \ref{sec:unsup} and \ref{sec:bert} report a discussion of, respectively, the supervised and unsupervised approaches. An empirical validation is in Section \ref{sec:exp}. Conclusions and outlooks are finally reported in Section \ref{sec:conc}.

\section{Literature Review}\label{sec:existing}
The onset of DL has given drive to powerful data processing models, which facilitate NLP applications. In this context, systems that understand the semantic and syntactic aspects of a text are extremely important. Word embedding models such as \emph{Word2Vec} \cite{mikolov}, or \emph{Glove} \cite{pennington2014glove}, as well as sentence embedding models such as USE (\emph{Universal Sentence Encoder}, \cite{cer2018universal}) help in representing text as a mathematical object in a reliable way. The text representations using such embeddings along with NLP and deep neural networks \cite{hochreiter1997long,mou2016transferable} played a vital role in text extraction, classification and clustering.

Another major shift was the introduction of attention models and transformers \cite{vaswani2017attention}, these are language models able to better understand text semantics by contextual analysis. BERT, ELMO \cite{peters2018deep} and various versions of these models gave a big boost to recent NLP applications \cite{hassan2019bert}. Moreover, word embeddings, originally intended as static model of a given corpus, later led to the exploration of their dynamic evolution over time, this being mainly used to compare the semantic shifts of words over time and detection of word analogies \cite{kulkarni2015statistically,kutuzov2018diachronic}. Notably, some of these works used BERT for story ending predictions and temporal event extractions \cite{han2019contextualized,li2019story}. In the next sections, we show how these models can be applied to literary text understanding.

\section{Character Trajectories by Temporal Word Embeddings}\label{sec:unsup}
Both the unsupervised technique presented in this section and the supervised approach discussed in Section \ref{sec:bert} require a reliable identification of the characters involved in the plot. This corresponds to a named entity recognition task, for which standard tools can be used.\footnote{E.g., see \url{https://nlp.stanford.edu/software/CRF-NER.html}.} As same characters can occur in the text with different aliases (e.g., \emph{Ron} and \emph{Ronald} Weasley), a \emph{de-aliasing} might be needed as an additional pre-processing step. We achieve that by a clustering of the named entities based on the DBSCAN algorithm \cite{DBSCAN}. The entities are clustered using precomputed distances based on the \emph{sequence matcher} algorithm, which finds the longest common subsequences.

After character identification and de-aliasing, learning the embeddings of the characters of a literary text is a straightforward task. As the learning of an embedding is based on contextual information, the only important condition is that a sufficient amount of co-occurrences of the characters in the text is available. If this is the case, the relative distances between the vectors can be used as proxy indicators of the relations between the corresponding characters. This can be also achieved for separate parts of a same text (e.g., chapters), provided that the amount of text remains sufficient for learning. In this way it is possible to capture the relations between characters for each part, but not to describe the dynamic evolution of the same character over the whole text. Vectors trained in different embeddings, even with the same dimensionality, are in fact not directly comparable.

The method employed in \cite{di2019training} elegantly addresses this issue by \emph{aligning} different temporal representations using a shared coordinate system. The model uses a skip-gram Word2Vec architecture, where the context matrix (the output weight matrix) is fixed during the training, while allowing the word embedding input weight matrices to change on the basis of co-occurrence frequencies that are specific to a given temporal interval. After training, model returns the context embeddings, that we are going to consider as a \emph{temporal} word embedding. To achieve that, first, a static word embedding is trained with random initialization using the whole text and ignoring temporal slices.

\begin{figure}[htp!]
\centering
\includegraphics[width=0.6\linewidth]{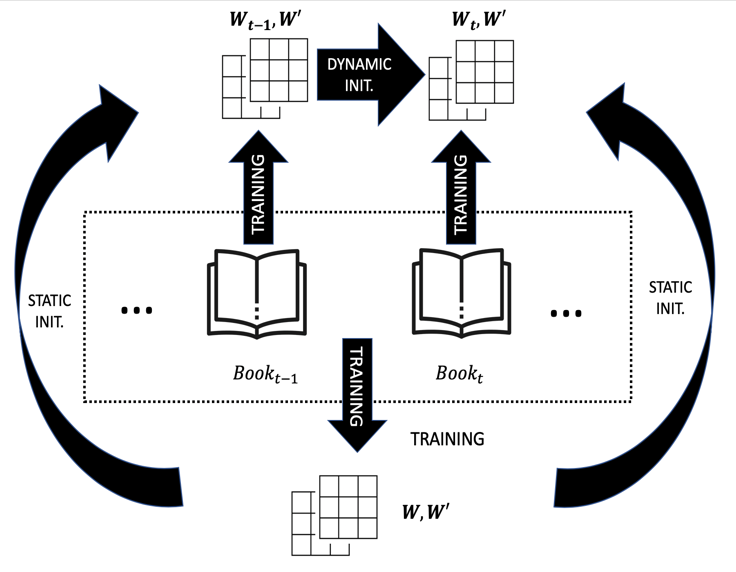}
\caption{Training temporal embeddings}
\label{fig:dynamic}
\end{figure}

Let us denote as $\bm{W}$ the corresponding word embedding matrix and as $\bm{W}'$ the corresponding context matrix. For each slice, we instead initialize the word embedding matrix with $\bm{W}$ while keeping $\bm{W}'$ as a frozen context matrix equal for all the time slices \cite{di2019training}. This initialization has been proved to force alignment and make it possible to compare vectors from embeddings associated to different time slices. The architecture is depicted in Figure \ref{fig:dynamic}. In particular, we adopt the dynamic initialization scheme proposed in \cite{volpettitemporal}, which appears to be more suitable for narrative text because of its intrinsic sequential nature.

Dynamic embeddings, generally used for word analogies, are considered here to describe and interpret relations by means of the trajectories spanned by the vectors associated to different characters. The character embeddings are represented in a visual space by dimensionality reduction \cite{maaten2008visualizing} to understand the evolving relations between characters, in terms of time slices in a novel such as chapters or other parts of the text. This could be further related to character sentiments, clustering of emotions and other descriptions.

\section{BERT-based Classification of Character Relations}\label{sec:bert}
The unsupervised approach considered in the previous section describes the relation between characters in terms of relative positions of the corresponding vectors and their evolution. Here we consider a character relation extraction based on binary classifiers. This is a supervised approach based on a ground truth of annotated sentences where the two characters are identified together with a Boolean value expressing whether or not the relation under consideration is met. The character names or aliases are eventually replaced by anonymous placeholders, as this helps the model to learn the relationships by abstracting from the specific names.

For the learning phase, we use the BERT classification model. Its pre-trained model can be fine tuned for classification with an additional output layer. BERT has a wordpiece tokenizer using two special tokens (SEP and CLS), which are used to encode valuable information of sentence structure and semantics after fine-tuning. The BERT-base has twelve transformer layers and in the classification task, the pooled token embedding from the CLS tokens is fed into a linear classifier for predictions. With the powerful attention mechanisms, BERT embeddings encode deep semantic and syntactic contextual information. The relation extraction problem is modelled as a single sentence classification task using BERT model. More details about this general architecture are in \cite{devlin2019bert}.

As creating ground truth in this field might be very expensive, we also discuss techniques for automatic data annotation. As an example, let us focus on family relations, where the problem is to decide whether or not there is familial relation between two characters. By increasing the neighbourhood parameter, the output of the DBSCAN clustering algorithm used for de-aliasing produces clusters in which the characters belonging to same family are together (as their second names remain same). E.g., in the \emph{Harry Potter} books, we have clusters with the \emph{Potter} and the \emph{Weasley} family. These clusters are used for automatic creation of the positive samples in training data, while the remaining characters are used for negative sample generation.

\section{Experimental Analysis}\label{sec:exp}
The above discussed approaches to character relation modelling (Section \ref{sec:unsup}) and understanding (Section \ref{sec:bert}) are validated here with two novels: \emph{Little Women} (LW) by L.M. Alcott (text length $197`524$ words) and the first six books of the \emph{Harry Potter} series (HP) by J.K. Rowling ($885`943$ words).

\paragraph{\bf Family Relationship Classification.} Due to the intricate nature of its plot and its length, HP is being often used as a benchmark for natural language understanding in literary domain \cite{bonato2016mining,sparavigna2013social}. As an application of the ideas discussed in Section \ref{sec:bert}, we consider the task of predicting whether or not a given pair of characters has a family relation or not. A BERT based classifier is used for that.\footnote{See \url{https://github.com/huggingface/transformers}.} Out of six books, we use the sentences generated from five books as training set and the remaining book as a test according to a cross-validation scheme. For the training set, the automatic class labelling is done by creating clusters for the same family groups (see Section \ref{sec:bert}). 
The number of samples for each book is $160$, $250$, $239$, $396$, $478$, and $231$, the ratio of positive samples for each book being $30.0\%$,
$39.2\%$, $28.9\%$, $38.6\%$, $62.6\%$, and $47.6\%$. BERT is used together with the Adam optimizer \cite{kingma2014adam}. This gives a learning rate equal to $2 \cdot 10^{-5}$, warm up equal to $0.1$ and ten epochs. The results in Table \ref{tab:results} show reasonably good average performances and their standard deviations over the six books. Note that the aggregated values, corresponding to weighted averages, might be higher than those for negative or positive samples only.

\begin{table}[h!]
\begin{center}
\caption{Character familial relation classification in \emph{Harry Potter}}
\label{tab:results}
\begin{tabular}{lp{1cm}rp{1cm}rp{1cm}r}
\hline
Samples&&Precision&&Recall&&F-score\\
\hline
Negative&& $79\pm 13\%$&&$85\pm 7\%$&&$81\pm 8\%$\\
Positive&& $77\pm 9\%$&&$71\pm 9\%$&&$73\pm 4\%$\\
All&& $80\pm 4\%$&&$78\pm 7\%$&&$78\pm 7\%$\\
\hline
\end{tabular}
\end{center}
\end{table}

A test is also done on the LW benchmark with the HP training data. A lower F-score level ($64\%$) is obtained, possibly related to an over-fitting effected. This might be relevant for literary texts, where the differences between different data (e.g., different texts of different authors) are typically stronger than in other domains.

It is important to note we have implemented a classification model whose predictions are at the sentence level. When coping with character pairs, it would be more appropriate to consider a higher level, i.e, prediction with respect to all the sentences that express the character pair relation. This is achieve by a \emph{bag of sentence} approach, where a character pair is considered to have a relation, if at least one of the sentences is predicted as positive. For HP there are $85$ entity pairs ($12$ positive and $73$ negative) and the results are $9$ positive ($75\%$) and $66$ ($90\%$) negative pairs correctly predicted. Considering the intrinsic complexity of literary text, where sentences might have very complex structures, this might be regarded as a promising result and also advocate our strategy for the automatic generation of training set.

\begin{figure}[htp!]
\centering
\begin{tikzpicture}[scale=1.5]
\begin{axis}[legend pos=north west,legend style={font=\tiny},
ytick=\empty,
xticklabels={,{\tiny Book I},,{\tiny Book II},,{\tiny Book III},,{\tiny Book IV},,{\tiny Book V}},
xtick={0,0.5,1,1.5,2,2.5,3,3.5,4,4.5,5}]
\addplot[black!30] coordinates {(0.3333333333333333,0.055697430765069944)(0.5,0.00899106720841314)(0.6666666666666666,0.044493068618475684)(0.8333333333333333,0.013151304916594664)(1.0,0.011214634948915725)(1.0,0.06310487751123339)(1.1666666666666667,0.032442398095156366)(1.3333333333333333,0.015389980849714102)(1.5,0.04461684162767632)(1.6666666666666665,0.037125010658615065)(1.8333333333333333,0.046067112601006044)(2.0,0.08028293094346228)(2.0,0.07157398330428943)(2.125,0.049529263485158515)(2.25,0.04027532399859457)(2.375,0.03358789754846936)(2.5,0.044683256185126696)(2.625,0.025546839797264242)(2.75,0.04699407180241011)(2.875,0.012993036066271446)(3.0,0.008721775580141533)(3.0,0.10286542543454968)(3.0588235294117645,0.08821159099765541)(3.1176470588235294,0.036035022159017105)(3.176470588235294,0.033785605421382314)(3.235294117647059,0.023903160182944916)(3.2941176470588234,0.0629390884090626)(3.3529411764705883,0.04923096586348452)(3.4705882352941178,0.056501794082085666)(3.5294117647058822,0.012749420184309246)(3.588235294117647,0.061756936604156265)(3.6470588235294117,0.02832917337646701)(3.7058823529411766,0.031237836733785773)(3.764705882352941,0.0760782655624912)(3.8235294117647056,0.03517947362281404)(3.8823529411764706,0.029532597333841726)(3.9411764705882355,0.07603127949173472)(4.0,0.02709436042440927)(4.0,0.11422836728115504)(4.083333333333333,0.07886008282071744)(4.166666666666667,0.12411137542713602)(4.25,0.061446681205068576)(4.333333333333333,0.05194004608174596)(4.416666666666667,0.080663351371148)(4.5,0.041306961243205986)(4.583333333333333,0.03044873786965352)(4.666666666666667,0.0833190402255618)(4.75,0.06635781688370002)(4.833333333333333,0.11421493965838547)(4.916666666666667,0.17694407917295363)(5.0,0.095970115329991)};
\addplot[] coordinates {(0.16666666666666666,0.054023700241486394)(0.3333333333333333,0.01588680608257753)(0.5,0.032340175971085516)(0.6666666666666666,0.035002456329140896)(0.8333333333333333,0.011726632859257347)(1.0,0.009997593531240856)(1.0,0.05717161051842401)(1.1666666666666667,0.02016461768677036)(1.3333333333333333,0.010412291664277262)(1.5,0.06372886014026458)(1.6666666666666665,0.037064767552647204)(1.8333333333333333,0.00959967735340439)(2.0,0.022003732142110377)(2.0,0.07045918629835157)(2.125,0.032080477614050085)(2.25,0.020566935586175616)(2.375,0.035061270066559236)(2.5,0.043463699723881644)(2.625,0.03369593715819852)(2.75,0.027017417627986817)(2.875,0.017428381653430636)(3.0,0.021266521631402635)(3.0,0.1036071508987898)(3.0588235294117645,0.04954374392930716)(3.1176470588235294,0.022990256782397744)(3.176470588235294,0.03164579260891742)(3.235294117647059,0.020527025441055446)(3.2941176470588234,0.05507505652043965)(3.3529411764705883,0.04933253420741579)(3.4705882352941178,0.02601916740607413)(3.5294117647058822,0.004125391458897232)(3.588235294117647,0.029196458749644383)(3.6470588235294117,0.03121571412971902)(3.7058823529411766,0.021571994639028924)(3.764705882352941,0.04349526512755153)(3.8235294117647056,0.05558943931957172)(3.8823529411764706,0.03952948681984547)(3.9411764705882355,0.016855949358024258)(4.0,0.03138126616378667)(4.0,0.12750560873138683)(4.083333333333333,0.05458569344652242)(4.166666666666667,0.09782826004279632)(4.25,0.08076871476490366)(4.333333333333333,0.07487238916590988)(4.416666666666667,0.0576133830637402)(4.5,0.05289239570982529)(4.583333333333333,0.03218827444906702)(4.666666666666667,0.094801246426509)(4.75,0.021929130537898134)(4.833333333333333,0.11874981309925858)(4.916666666666667,0.1340155328891317)(5.0,0.09206238183812931)};
\addplot[dotted] coordinates { 
(0.0,0.03936666516330389)(0.8333333333333333,0.07028986024765016)
(1.0,0.013801808689667405)(1.0,0.052813533093232556)(1.8333333333333333,0.09797217399434976)(2.0,0.1367589107693713)(2.0,0.05481189653527829)(2.125,0.10868770237726721)(2.25,0.13240315588213747)(2.375,0.0556345618779398)(2.5,0.055714351707065624)(2.75,0.13436439787045895)(2.875,0.09984088115365786)(3.0,0.0635779192635324)(3.0,0.03638845556261139)(3.176470588235294,0.023288414535307722)(3.235294117647059,0.04197497185971755)(3.2941176470588234,0.08198705677004836)(3.3529411764705883,0.031130171549768715)(3.411764705882353,0.020772449964888584)(3.4705882352941178,0.037555154924698186)(3.5294117647058822,0.031074608580872742)(3.7058823529411766,0.07152508107258093)(3.8235294117647056,0.06588264940296307)(3.8823529411764706,0.0948769916056511)(4.0,0.06553653663811776)(4.083333333333333,0.19278779597040685)(4.166666666666667,0.1599901026137759)(4.25,0.09289990136989035)(4.333333333333333,0.17565624345163033)(4.416666666666667,0.18290546857241985)(4.5,0.12048023788514262)(4.583333333333333,0.12510629346171964)(4.666666666666667,0.19571999158788644)(4.75,0.16597809025731391)(4.833333333333333,0.07441552436072385)(4.916666666666667,0.1860592798438533)};
\legend{Hermione,Ron,Voldemort}
\end{axis}
\end{tikzpicture}
\caption{Characters trajectories for \emph{Harry Potter}
\label{fig:book5hp4}}
\end{figure}
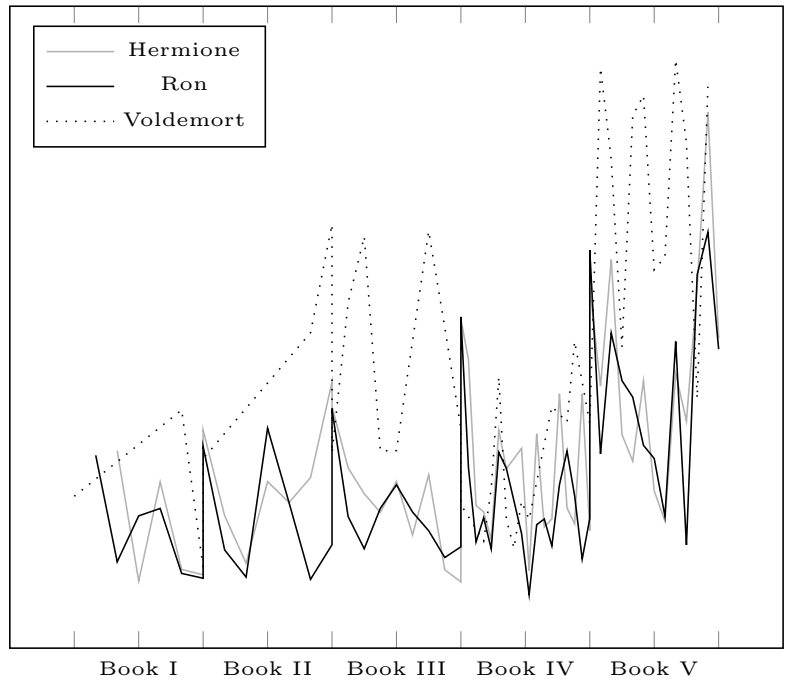

\paragraph{\bf Temporal Word Embeddings.}
Following the discussion in Section \ref{sec:unsup}, we train a temporal word embedding for the first six books of HP. We focus on the four characters which appear more frequently. The static embedding is trained with the whole text of each book, while the dynamic embeddings are based on sub-slices containing text of length equal to $1000$ characters. For each character, we extract the corresponding trajectory for each book. For a better interpretation of the relations, we consider the main character (i.e., \emph{Harry}) and plot the evolution over time of the relative (cosine) distances from the other characters. Since these vectors embed semantic information, it is expected that in the trajectories corresponding to smaller distances correspond to closer relations with Henry. In fact, the results in Figure \ref{fig:book5hp4} show that the trajectories of positive characters or friends (i.e., \emph{Ron} and \emph{Hermione}) move in a similar way. The main antagonist (i.e., \emph{Voldemort}) is found instead to move in a different direction and at a higher distance.

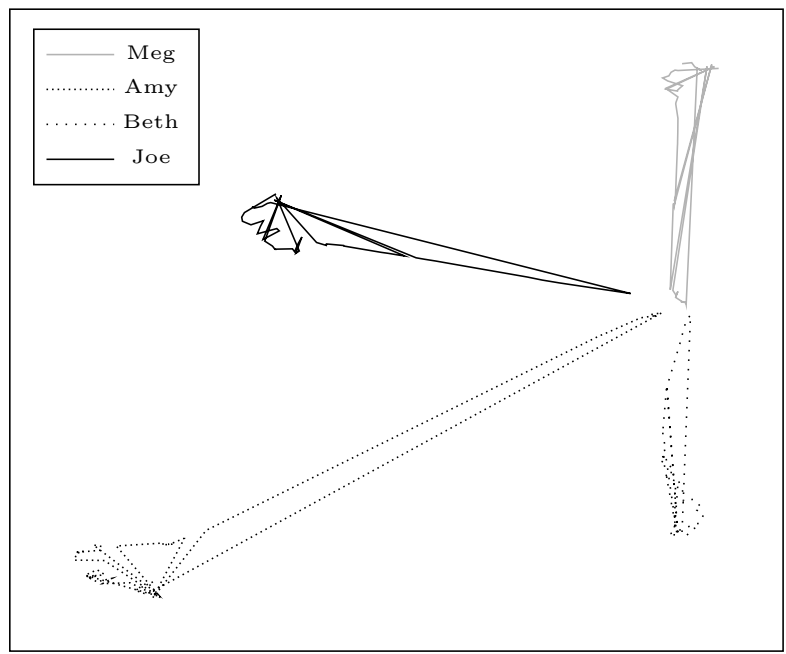
\begin{figure}[htp!]
\begin{tikzpicture}[scale=1.5]
\pgfplotsset{ticks=none}
\begin{axis}[legend pos=north west,legend style={font=\tiny}]
\addplot [black!30] coordinates {(4.5099607,9.145459)(3.57413,9.0662775)(3.5041938,9.081636)(3.4137459,9.000664)(3.363038,8.866242)(3.2196476,8.789564)(3.336514,8.677338)(3.67762,8.499175)(3.5649667,8.337997)(3.5615878,8.298908)(3.3058891,8.368536)(4.411359,9.214104)(3.2917495,8.400075)(3.572686,8.079635)(3.5276706,7.8792367)(3.5218406,7.819681)(3.5725102,7.283179)(3.5694788,6.764239)(3.564716,6.409861)(3.5642204,6.320786)(3.5614207,6.2602897)(3.5108786,4.1908727)(4.3509064,9.292666)(3.4946973,4.240342)(3.489429,4.2161703)(3.5081418,4.156599)(3.4836624,3.9927704)(3.4858332,3.9810376)(3.4876406,4.0152354)(3.4760244,4.1220536)(3.48372,4.062978)(3.4633827,4.0681324)(3.3955011,0.88099146)(4.241013,9.220942)(3.4554677,0.8562715)(3.530878,0.6703262)(3.5720026,0.8179651)(3.519464,0.59453213)(3.6843014,0.42055798)(3.6399596,0.4366883)(3.7500453,0.424067)(3.7618566,0.33342305)(4.0158863,9.151562)(4.115068,9.066803)(3.9582112,9.206522)(3.8896604,9.35884)(3.6682746,9.320783)};
\addplot [densely dotted] coordinates {(-8.342304,-10.610808)(-9.04275,-10.060997)(-9.673887,-10.093517)(-9.451507,-9.894799)(-9.805856,-10.021437)(-10.046675,-9.8279)(-9.802068,-9.778855)(-10.02607,-9.885238)(-9.633703,-9.747921)(-9.800576,-9.574567)(-10.076173,-9.779362)(-8.397992,-10.55506)(-9.631561,-9.232356)(-9.934353,-9.219695)(-10.326742,-9.217409)(-10.299923,-9.105268)(-10.272848,-8.962244)(-10.114175,-8.881637)(-10.313372,-8.93368)(-9.729081,-8.846545)(-9.693886,-8.706198)(-9.88598,-8.648996)(-8.406389,-10.479309)(-9.358188,-8.688009)(-8.273186,-8.576298)(-8.297483,-8.667865)(-8.266495,-8.644202)(-8.0628805,-8.617887)(-8.164219,-8.62559)(-8.125448,-8.616445)(-8.125835,-8.620081)(-7.810387,-8.408485)(-7.792922,-8.411057)(-8.514099,-10.614926)(-7.28948,-8.103061)(-7.281028,-8.09644)(1.7779036,-0.83017725)(1.8701068,-0.77361786)(2.7121656,-0.16919155)(3.0169158,-0.09458717)(2.9432204,-0.047557563)(3.1993005,0.0181159)(-8.582332,-10.427661)(-8.679336,-10.478853)(-8.920881,-10.344939)(-8.824386,-10.267015)(-8.461559,-9.987007)};
\addplot [dotted] coordinates {(3.399134,-8.2278385)(3.9947112,-7.8537307)(4.061526,-7.8462396)(4.165823,-7.5131683)(4.0054226,-7.345303)(4.0395436,-7.191584)(4.0945888,-7.213176)(3.932565,-6.940311)(3.700814,-7.0141015)(3.6766925,-6.587878)(3.5778377,-6.2286468)(3.485731,-8.295013)(3.5197384,-6.2659173)(3.2941034,-6.1130967)(3.408607,-6.144191)(3.4197803,-5.8867893)(3.4845016,-5.838639)(3.4017882,-5.723489)(3.138255,-5.4674745)(3.1901884,-5.452265)(3.2916765,-5.3035164)(3.211435,-5.235973)(3.5488815,-8.21253)(3.2862277,-5.2513647)(3.2602026,-5.086409)(3.2939625,-5.0978804)(3.2242289,-4.5384245)(3.2285585,-4.3551507)(3.2287626,-4.2848268)(3.227612,-4.3508224)(3.3147044,-2.6714005)(3.3094282,-2.7399797)(3.314358,-2.743133)(3.5742745,-8.321654)(3.3214943,-2.7259347)(3.615556,-1.143696)(3.6386466,-0.9926742)(3.74747,-0.3511998)(3.859905,-0.05374667)(3.8619034,-0.062683135)(3.6643476,-8.259762)(3.773303,-8.295963)(3.9048634,-8.060958)(4.026101,-7.9769216)(4.0504045,-7.907703)};
\addplot [black] coordinates {(-5.6824794,4.460025)(-6.4085937,3.7616632)(-6.4715667,3.5910206)(-6.4541655,3.4378214)(-6.258584,3.3053043)(-5.971511,3.4570346)(-6.040031,3.2492507)(-6.1116266,2.9683256)(-5.6629906,3.1651464)(-5.606908,3.0784843)(-5.9818015,2.7563956)(-5.560352,4.4049788)(-5.9407244,2.6969256)(-5.762161,2.5101566)(-5.707371,2.3922527)(-5.725791,2.3947744)(-5.301888,2.408071)(-5.2534933,2.301176)(-5.171372,2.6046357)(-5.0860906,2.849026)(-5.225759,2.2410443)(-5.1457667,2.3276696)(-5.6619635,4.303816)(-4.7480364,2.6451597)(-4.52371,2.5354176)(-4.51918,2.5908763)(-4.123952,2.5323386)(-4.1186476,2.5123463)(-3.6114733,2.3727474)(-3.2577467,2.2763653)(-3.0931826,2.2395287)(-2.8423347,2.1711357)(-2.7230377,2.1433647)(-5.724335,4.2288923)(-2.4566236,2.0735948)(-2.2592647,2.0183697)(-2.0057557,1.9534389)(-1.5673388,1.8283334)(0.11418169,1.3523777)(0.44578424,1.2423322)(0.83989626,1.1366738)(2.4818993,0.7392258)(-5.8047485,4.1351876)(-5.882184,4.1038823)(-6.0010815,3.994064)(-6.180232,3.9289753)(-6.236088,3.9846852)};
\legend{Meg,Amy,Beth,Joe}
\end{axis}
\end{tikzpicture}
\caption{Character trajectories for \emph{Little Women}\label{fig:lw}}
\end{figure}

A similar analysis for LW is reported in Figure \ref{fig:lw}. In this case we display a t-SNE \cite{maaten2008visualizing} two-dimensional projection of the vectors over different groups of chapters for the four major characters (i.e., the four \emph{March} sisters). As a comment, the temporal word embedding seems to capture the separation, during the central part of the plot, between the characters of \emph{Joe} and \emph{Amy}, i.e., the two characters who left their home town, and the other two sisters.

\section{Conclusion}\label{sec:conc}
In this paper, we presented supervised and unsupervised DL models for analysing and interpreting character relations in a novel. We used BERT classifiers for predicting the character relations, while an unsupervised approach based on temporal word embeddings was used to interpret the character relation evolution. Both methods are found to be promising to explore the relations involved within characters in a novel. Thus, the approaches can be further applied to literary text understanding for deriving character networks and hence studying the relations and sentiments involved. In future, we want to integrate these approaches to build a more user-friendly tool to analyse the character networks and use it for an extensive validation.

\bibliographystyle{splncs04}
\bibliography{biblio}
\end{document}